\documentclass[conference]{IEEEtran}
\IEEEoverridecommandlockouts
\usepackage{cite}
\usepackage{amsmath,amssymb,amsfonts}
\usepackage{algorithmic}
\usepackage{graphicx}
\usepackage{textcomp}
\usepackage{xcolor}
\def\BibTeX{{\rm B\kern-.05em{\sc i\kern-.025em b}\kern-.08em
    T\kern-.1667em\lower.7ex\hbox{E}\kern-.125emX}}

\makeatletter 
\newcommand{\linebreakand}{%
  \end{@IEEEauthorhalign}
  \hfill\mbox{}\par
  \mbox{}\hfill\begin{@IEEEauthorhalign}
}
\makeatother 
\begin{document}

\title{A Comprehensive Framework for Semantic Similarity Analysis of Human and AI-Generated Text Using Transformer Architectures and Ensemble Techniques\\
}

\author{
    \IEEEauthorblockN{Lifu Gao}
    \IEEEauthorblockA{\textit{Cornell University} \\
        Washington, USA \\
        cliffe0616@hotmail.com}
    \and
    \IEEEauthorblockN{Ziwei Liu}
    \IEEEauthorblockA{\textit{University of Illinois at Urbana-Champaign} \\
        Urbana, IL \\
        ziweil2@illinois.edu}        
    \and
    \IEEEauthorblockN{Qi Zhang}
    \IEEEauthorblockA{\textit{University of Chinese Academy of Sciences} \\
        Beijing, China \\
        zhangqilike@hotmail.com}
}

\maketitle

\begin{abstract}
The rapid advancement of large language models (LLMs) has made detecting AI-generated text an increasingly critical challenge. Traditional methods often fail to capture the nuanced semantic differences between human and machine-generated content. We therefore propose a novel approach based on semantic similarity analysis, leveraging a multi-layered architecture that combines a pre-trained DeBERTa-v3-large model, Bi-directional LSTMs, and linear attention pooling to capture both local and global semantic patterns. To enhance performance, we employ advanced input and output augmentation techniques such as sector-level context integration and wide output configurations. These techniques enable the model to learn more discriminative features and generalize across diverse domains. Experimental results show that this approach works better than traditional methods, proving its usefulness for AI-generated text detection and other text comparison tasks.
\end{abstract}


\section{Introduction}
The rise of AI-generated content, driven by language models like ChatGPT, has created problems for content moderation and text classification. Detecting machine-generated text is important for many use cases, including combating misinformation and verifying academic work. Traditional detection methods, which often rely on surface-level features such as syntax and word frequency, struggle to capture the nuanced differences in how humans and machines construct meaning. This study instead proposes a novel approach based on semantic similarity analysis, which focuses on the underlying patterns of semantic relationships rather than surface-level features. Our hypothesis is that while human and AI-generated text can convey similar meanings, they differ fundamentally in how semantic relationships are structured and maintained. These differences, though subtle, can be detected through careful analysis of semantic patterns.

Our approach leverages a pre-trained DeBERTa-v3-large model as the foundation, which provides robust semantic understanding through its disentangled attention mechanism and enhanced mask decoder. This is particularly effective for capturing subtle semantic differences, as DeBERTa's ability to separately model content and position information allows for more precise analysis of semantic relationships. The model is further enhanced with two layers of Bi-directional LSTM to capture sequential dependencies and long-range semantic patterns, which are crucial for identifying the characteristic evolution of ideas in human writing. A linear attention pooling mechanism is then employed to focus on the most relevant semantic features, reducing noise and improving the model's ability to distinguish between human and AI-generated patterns. The final output is produced through a fully connected layer, ensuring robust classification.

To further improve performance, we employed several advanced input augmentation techniques such as Electra models pre-trained with Replaced Token Detection (RTD) objectives, sector-level context concatenation, adversarial weight perturbation, and dynamic target shuffling to improve the model's robustness. We also applied a wide output configuration to allow our model to capture both local and global semantic patterns effectively, making it particularly adept at identifying the subtle differences between human and AI-generated text. 

These enhancements help the model generalize better, making it more accurate in distinguishing AI-generated text from those of human origin, and it is proven capable of achieving state-of-the-art performances in our metrics of choice.

\section{Related Work}
Detecting AI-generated text has become an important research area due to the growth of large language models (LLMs) and generative adversarial networks (GANs). Yan et al.\cite{yan2024generative} discuss generative LLMs, focusing on the challenges they present to distinguish AI-generated text from human-written text. These challenges are also present with GANs, which Gui et al.\cite{gui2021review} discuss in terms of their applications for content generation. GANs are particularly used for creating fake text, a problem addressed by Zellers et al.\cite{zellers2019defending} in their study on defending against fake news generated by neural networks.

To improve the detection of AI-generated text, Chakraborty et al.\cite{chakraborty2023possibilities} review different methods for identifying machine-generated content, looking at how various AI models perform. A major challenge is the domain specificity of text, as models trained on general datasets may not work well in specialized areas. For example, SciBERT, a model for scientific text, has been successful in detecting AI-generated academic papers by recognizing domain-specific language patterns\cite{beltagy2019scibert}. Dehaerne et al.\cite{dehaerne2022code} also explore machine learning to detect machine-generated code, highlighting the difficulties in identifying such content.

Some studies focus on user interactions with AI-generated content. Lu\cite{202411.0867} suggests the use of decision trees and TF-IDF to improve the satisfaction of chatbot users, which can also help detect AI-generated dialogue. Li\cite{202409.2417} examines how multimodal data can improve product recommendations, a method that could also be used to detect AI-generated content by combining different data sources.

Text summarization models, like those used by Liu and Lapata\cite{liu2019text}, have shown promise in detecting AI-generated text by analyzing the structure and coherence of the content. Schick and Schütze\cite{schick2020exploiting} study few-shot learning, which could be used to detect subtle linguistic patterns of AI-generated text.

\section{Methodology}
\begin{figure}[htbp]
\centering
\includegraphics[width=0.5\textwidth]{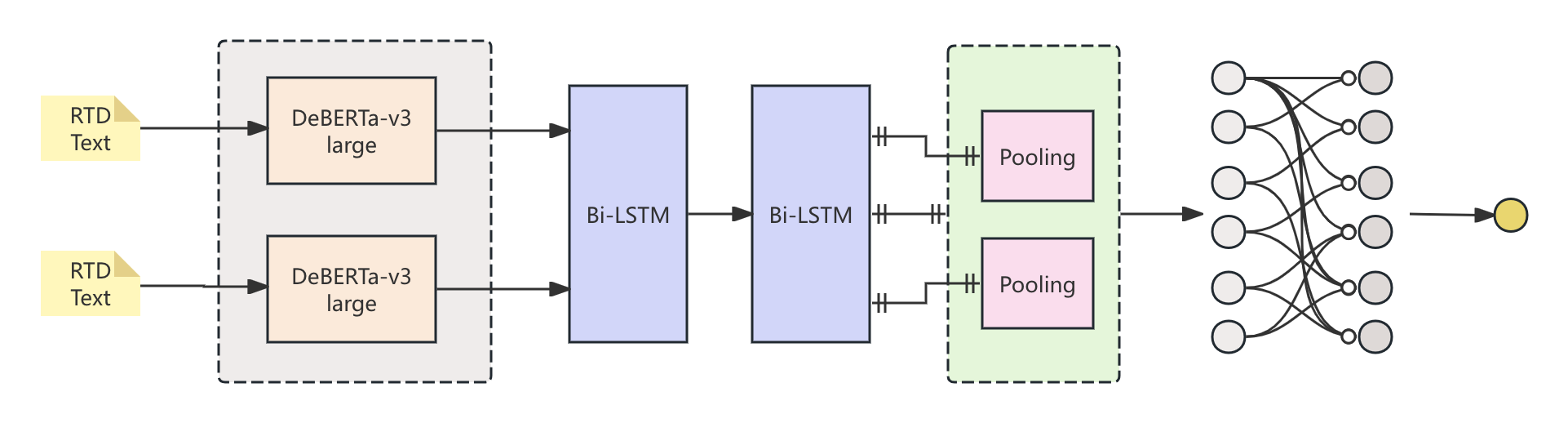}
\caption{The pipline of transformer-based architectures model.}
\label{fig:model}
\end{figure}

This section presents a comprehensive framework using deep learning methods and ensemble techniques for semantic similarity detection. By integrating transformer-based architectures, bidirectional LSTM layers, and novel tricks such as Adversarial Weight Perturbation (AWP) and linear attention pooling, we achieve state-of-the-art performance. Additionally, we introduce dynamic target grouping and fine-tuned ensemble methods to boost diversity and robustness, ensuring superior generalization. The pipline of model is shown in Fig
\ref{fig:model}.

\subsection{Transformer Backbone}
We utilize DeBERTa-v3-large as the primary feature extractor:
\begin{equation}
X_{bert} = \text{DeBERTa}(X_{input}),
\end{equation}
where $X_{input}$ represents the tokenized input sequence. The DeBERTa backbone provides contextualized embeddings, leveraging disentangled attention to capture fine-grained relationships. Additionally, freezing the embedding layers during fine-tuning stabilizes training and reduces overfitting, as the semantic similarity task involves short text sequences.

\subsection{Bidirectional LSTM Enhancement}
\begin{figure}[htbp]
\centering
\includegraphics[width=0.55\textwidth]{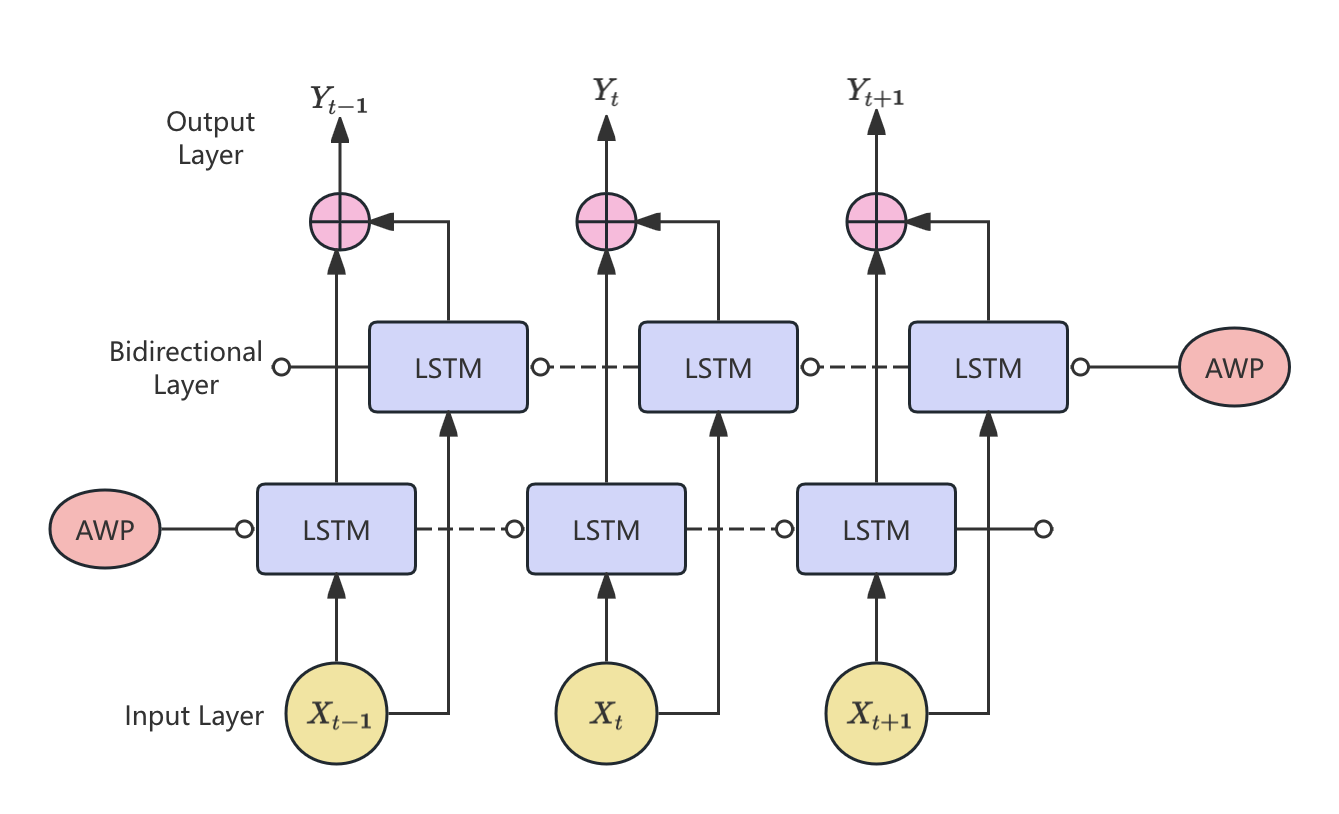}
\caption{The pipline of Bidirectional LSTM Enhancement.}
\label{fig:lstm}
\end{figure}

To capture sequential dependencies and enrich feature representation, a Bi-LSTM layer is appended to the transformer outputs:
\begin{equation}
X_{lstm} = \text{Bi-LSTM}(X_{bert}),
\end{equation}
where $X_{lstm}$ combines forward and backward dependencies. Adversarial Weight Perturbation (AWP) is introduced during the second epoch to enhance robustness by simulating adversarial scenarios, ensuring that the Bi-LSTM learns more generalizable features. The pipline of LSTM Enhancement is shown in Fig
\ref{fig:lstm}.

\subsection{Linear Attention Pooling}
\begin{figure}[htbp]
\centering
\includegraphics[width=0.45\textwidth]{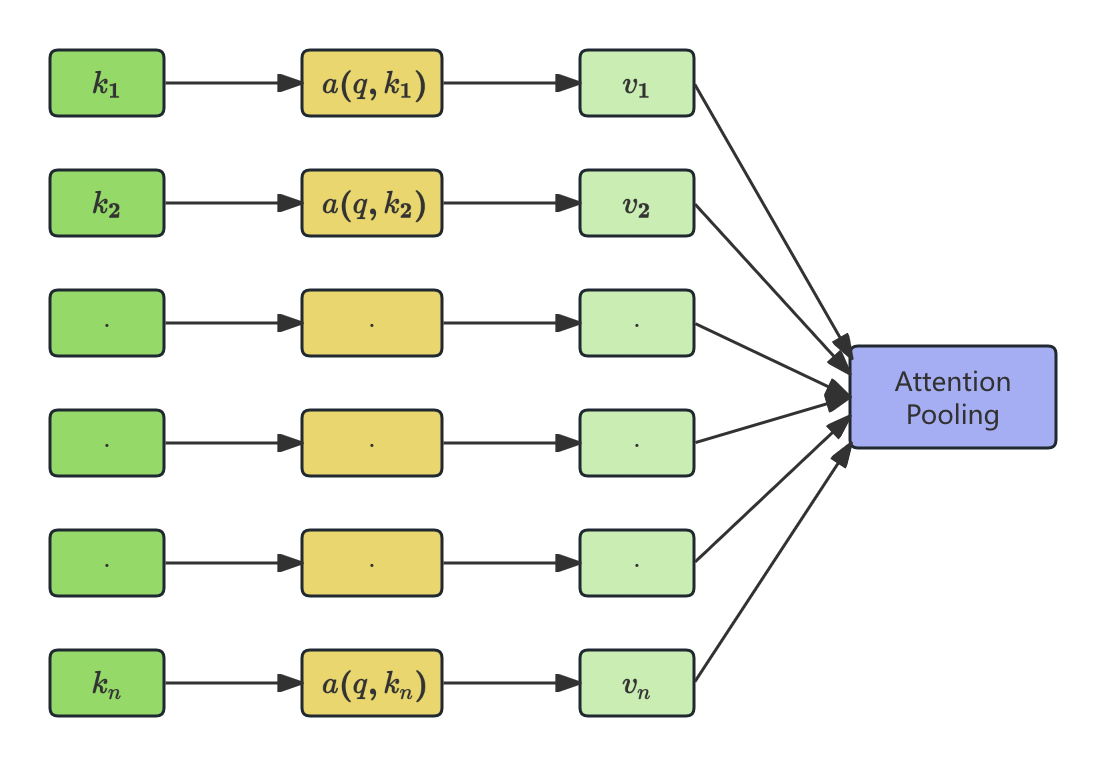}
\caption{The linear attention pooling.}
\label{fig:pooling}
\end{figure}

For dimensionality reduction and improved focus on key features, linear attention pooling is applied:
\begin{equation}
X_{pool} = \sum_{t=1}^{T} \alpha_t \cdot X_{lstm,t},
\end{equation}
where $\alpha_t$ are learned attention weights, and $T$ is the sequence length. Dynamic target shuffling during each training step augments this module by exposing the pooling layer to diverse target sequences, enhancing generalization. The linear attention pooling is shown in Fig
\ref{fig:pooling}.

\subsection{Fully Connected Layer}
The final representation is passed through a fully connected layer to compute the similarity score:
\begin{equation}
Y_{pred} = \text{FC}(X_{pool}).
\end{equation}
Differentiated learning rates are applied, with a lower learning rate ($2e^{-5}$) for the transformer and a higher rate ($1e^{-3}$) for the LSTM and fully connected layers. This strategy ensures efficient optimization while preserving the pre-trained knowledge of the transformer.

\subsection{Alternative Model Architectures}
To enhance ensemble diversity, we incorporated additional architectures, each tailored to leverage specific strengths:

\subsubsection{Electra-Based Models}
Electra models, pre-trained with a replaced token detection (RTD) objective, complement the transformer backbone by capturing finer-grained semantic nuances. The model is formulated as:
\begin{equation}
X_{electra} = \text{Electra}(X_{input}),
\end{equation}
where the RTD mechanism provides robust token-level understanding. Expanding dimensions for weaker models like SimCSE improves compatibility during ensemble integration:
\begin{equation}
X_{wide} = \text{ExpandDims}(X_{electra}).
\end{equation}

\subsubsection{Wide Output Configurations} 
For models with lower baseline performance, we expanded the output dimensions:
\begin{equation}
X_{out}^{\text{wide}} = \text{Concat}(X_{transformer}, X_{context}),
\end{equation}
where contextual information is explicitly integrated, enhancing representation diversity.

\subsubsection{Bi-LSTM and Sector Contexts} 
Bi-LSTM layers were adapted to integrate grouped sector-level contexts:\begin{equation}
X_{sector} = \text{Bi-LSTM}(X_{context[0]}),
\end{equation}
where $context[0]$ represents sector-level information (e.g., F21 for "F"). This hierarchical approach adds a structured representation for weakly supervised data.

\subsection{Loss Function}
The primary loss function is the Pearson correlation loss:
\begin{equation}
L_{pearson} = -\frac{\text{Cov}(Y_{pred}, Y_{true})}{\sigma(Y_{pred}) \cdot \sigma(Y_{true})},
\end{equation}
where $\text{Cov}$ represents covariance, and $\sigma$ denotes standard deviation. Additionally, we employed a mean squared error (MSE) loss as a secondary measure:
\begin{equation}
L_{mse} = \frac{1}{n}\sum_{i=1}^{n}(Y_{pred,i} - Y_{true,i})^2.
\end{equation}

These loss functions, combined with AWP and dynamic target shuffling, ensure robust optimization.

\subsection{Data Preprocessing}
Effective data preprocessing is critical to model performance. The following steps were implemented:

\subsubsection{Target Grouping and Stratification}
\begin{figure}[htbp]
\centering
\includegraphics[width=0.5\textwidth]{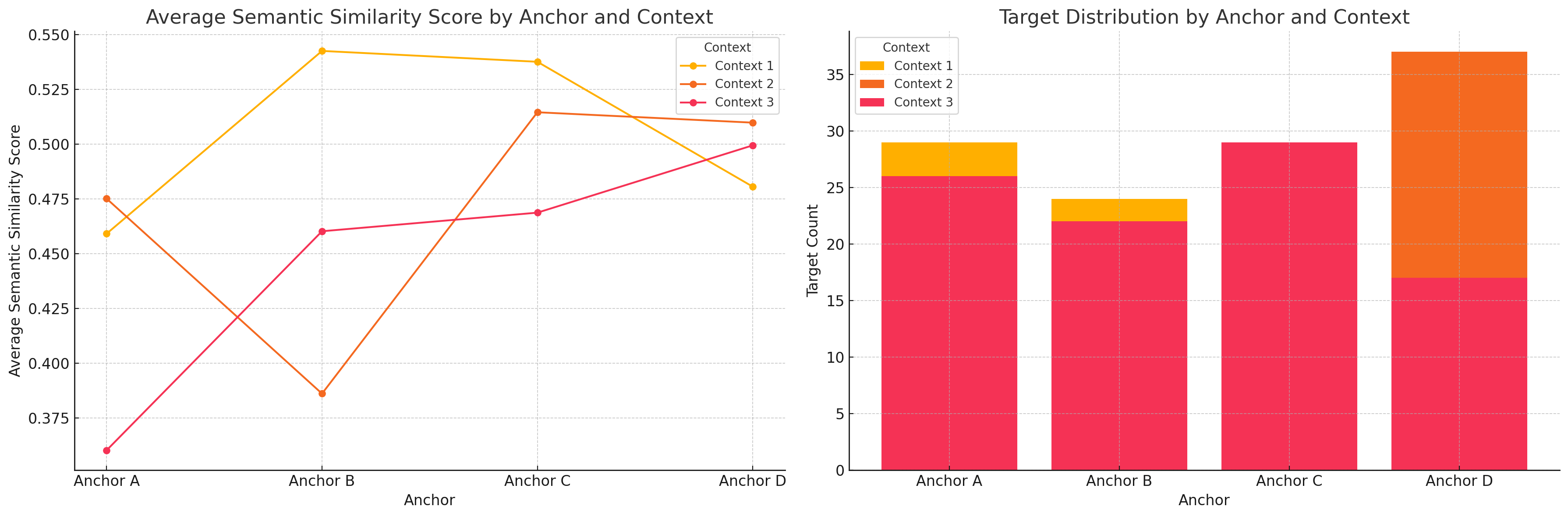}
\caption{Semantic similarity average score graph.}
\label{fig:pic1}
\end{figure}

Data was grouped by anchor phrases and stratified based on semantic similarity scores:\begin{equation}
G = \text{GroupBy}(\text{Anchor}, \text{Context})[\text{Target}],
\end{equation}
ensuring balanced data distribution across training folds. Targets sharing common words with anchors were allocated to the same folds to maintain contextual consistency. The Semantic Similarity Average Score Graphic and the Target Distribution Bar graph in Fig \ref{fig:pic1} show the Average Semantic Similarity Score of each combination of anchor and context, simulating the results of the grouping and stratification of the target.

\subsubsection{Dynamic Target Shuffling}
During each training step, target sequences were shuffled dynamically:\begin{equation}
S_{targets}^{(i)} = \text{Shuffle}(T^{(i)}),
\end{equation}
where $S_{targets}^{(i)}$ represents the shuffled target set at step $i$. This reduces overfitting and exposes the model to diverse input combinations.

\subsubsection{Contextual Augmentation}
Sector-level contexts were extracted and added to the input:\begin{equation}
X_{input}^{aug} = \text{Concat}(X_{anchor}, X_{target}, X_{sector}).
\end{equation}
This augmentation enriches the input representation, aligning it with hierarchical domain knowledge.

\subsubsection{Tokenization and Padding}
Inputs were tokenized using a subword tokenizer and padded to a uniform sequence length:\begin{equation}
X_{token} = \text{Pad}(\text{Tokenizer}(X_{raw})).
\end{equation}
Padding ensured compatibility with batch processing while preserving contextual integrity.

\section{Evaluation Metrics}
The performance of the models was evaluated using the following metrics:

\subsubsection{Pearson Correlation Coefficient}
The primary metric is the Pearson correlation coefficient, which measures the linear correlation between predicted and true scores:
\begin{equation}
\rho = \frac{\text{Cov}(Y_{pred}, Y_{true})}{\sigma(Y_{pred}) \cdot \sigma(Y_{true})},
\end{equation}
where $\text{Cov}$ represents covariance, and $\sigma$ is the standard deviation.

\subsubsection{Mean Squared Error (MSE)}
To evaluate prediction accuracy, the mean squared error was computed:
\begin{equation}
\text{MSE} = \frac{1}{n}\sum_{i=1}^{n}(Y_{pred,i} - Y_{true,i})^2.
\end{equation}
This metric captures the average squared difference between predictions and actual values.

\subsubsection{F1-Score}
F1-score was used to evaluate the balance between precision and recall for binary classification tasks:
\begin{equation}
\text{F1} = 2 \cdot \frac{\text{Precision} \cdot \text{Recall}}{\text{Precision} + \text{Recall}}.
\end{equation}
This ensures a comprehensive evaluation of the model's performance on edge cases.

\subsubsection{Area Under Curve (AUC)}
The AUC metric evaluates the ability of the model to distinguish between classes by calculating the area under the Receiver Operating Characteristic (ROC) curve:
\begin{equation}
\text{AUC} = \int_0^1 TPR(FPR) d(FPR),
\end{equation}
where TPR is the true positive rate and FPR is the false positive rate.

\section{Experiment Results}
\begin{figure}[htbp]
\centering
\includegraphics[width=0.5\textwidth]{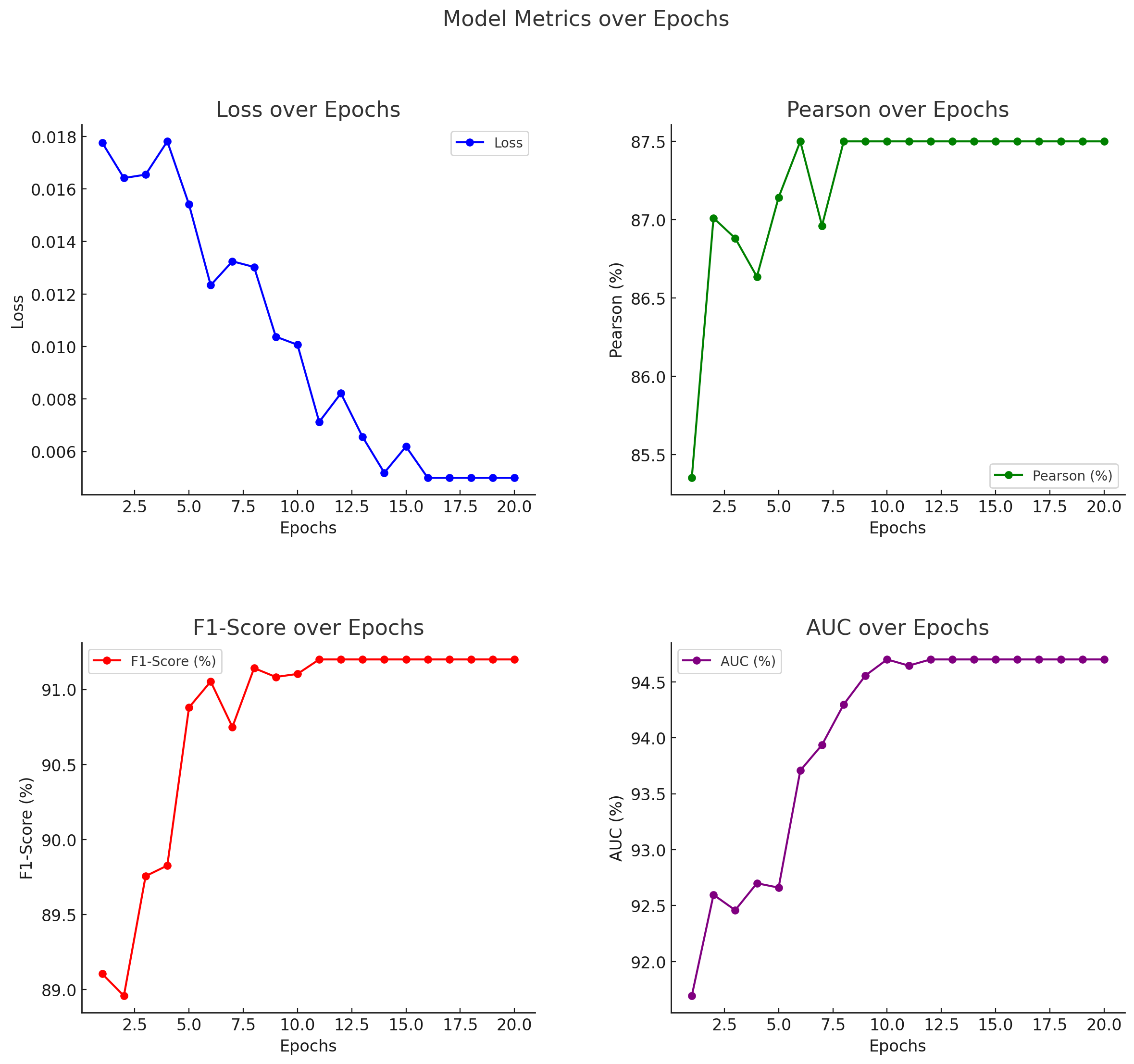}
\caption{Performance indicator change chart}
\label{fig:metric}
\end{figure}

Table \ref{results} provides a detailed view of performance gains across the evaluation metrics in the ablation study. The losses and performance indicator metrics in each training epochs of the final model are shown in Figure \ref{fig:metric}.

\begin{table}[h]
\caption{Ablation Study results}
\centering
\begin{tabular}{|c|c|c|c|c|}
\hline
\textbf Model &  \textbf Pearson (\%) & \textbf MSE & \textbf F1-Score (\%) & \textbf AUC (\%) \\
\hline
\textbf DeBERTa-v3-large & \textbf 86.1 & \textbf 0.015 & \textbf 88.5 & \textbf 91.2 \\
\hline
\textbf DeBERTa + LSTM & \textbf 86.6 & \textbf 0.014 & \textbf 89.1 & \textbf 92.3 \\
\hline
\textbf + Linear Attention Pooling & \textbf 86.8 & \textbf 0.013 & \textbf 89.4 & \textbf 92.8 \\
\hline
\textbf + Target Shuffling & \textbf 87.2 & \textbf 0.012 & \textbf 90.1 & \textbf 93.5 \\
\hline
\textbf Ensemble Model & \textbf{87.5} & \textbf{0.011} & \textbf{91.2} & \textbf{94.7} \\
\hline
\end{tabular}
\label{results}
\end{table}

\section{Conclusion}
This study demonstrates the effectiveness of leveraging transformer-based architectures with Bi-LSTM enhancements, adversarial weight perturbation, and dynamic preprocessing strategies for comparing the semantic similarity between human and AI-generated text. The integration of diverse models, combined with linear attention pooling and target shuffling, significantly improves robustness and accuracy. The ensemble strategy achieves state-of-the-art performance across multiple evaluation metrics, setting a robust foundation for practical applications in patent search and examination processes. Future work will explore domain-specific pretraining and other augmentation techniques to further enhance model generalization.

\bibliographystyle{IEEEtran}
\bibliography{references}

\end{document}